\documentclass{article}
\usepackage{ijcai23}

\usepackage{times}
\usepackage{soul}
\usepackage{url}
\usepackage[hidelinks]{hyperref}
\usepackage[utf8]{inputenc}
\usepackage[small]{caption}
\usepackage{amsmath}
\usepackage{color}
\usepackage{amsthm}
\usepackage{booktabs}
\usepackage{algorithm}
\usepackage{algorithmic}
\usepackage{multirow}
\usepackage{adjustbox}
\usepackage{arydshln}
\usepackage{bbding}
\usepackage{tikz}
\usepackage{amsmath}
\usepackage{bbding}
\usepackage{enumitem}
\usepackage{CJKutf8}

\usepackage{graphicx}
\usepackage{tikz-qtree}
\usepackage[edges]{forest}
\tikzset{%
    parent/.style = {align=center,text width=1.5cm,rounded corners=3pt, line width=0.3mm, fill=gray!10,draw=gray!80},
    background/.style = {align=center,text width=2cm,rounded corners=3pt, fill=blue!10,draw=blue!80,line width=0.1mm},   
    background_work/.style = {align=center, text width=5cm,rounded corners=3pt, fill=blue!10,draw=blue!0,line width=0.1mm},  
    datasets/.style = {align=center,text width=2cm,rounded corners=3pt, fill=red!10,draw=red!80,line width=0.1mm},
    datasets_work/.style = {align=center,text width=11cm,rounded corners=3pt, fill=red!10,draw=red!0,line width=0.1mm},
    challenges/.style = {align=center,text width=2cm,rounded corners=3pt, fill= cyan!10,draw= cyan!80,line width=0.1mm},
    challenges_work/.style = {align=center,text width=5cm,rounded corners=3pt, fill= cyan!10,draw= cyan!0,line width=0.1mm},
    methods/.style = {align=center,text width=2cm,rounded corners=3pt, fill= orange!10,draw= orange!80,line width=0.1mm},   
    methods_work/.style = {align=center,text width=6cm,rounded corners=3pt, fill= orange!10,draw= orange!40,line width=0.1mm},
    future/.style = {align=center,text width=2cm,rounded corners=3pt, fill= magenta!10,draw= magenta!80,line width=0.1mm},
    future_work/.style = {align=center,text width=5cm,rounded corners=3pt, fill= magenta!10,draw= magenta!0,line width=0.1mm},
}

\usepackage{xcolor}
\definecolor{myRed}{RGB}{204,65,37}

\usepackage{adjustbox}
\newcommand{\tablestyle}[2]{\setlength{\tabcolsep}{#1}\renewcommand{\arraystretch}{#2}\centering\footnotesize} 

\newcommand{\tf}[1]{\textbf{#1}}
\newcommand{\ti}[1]{\textit{#1}}
\newcommand{\tif}[1]{\textit{\textbf{#1}}}

\title{
    A Survey on Table-and-Text HybridQA: \\ Definitions, Methods, Challenges and Future Directions
}

\author{
    Dingzirui Wang, Longxu Dou, Wanxiang Che
    \affiliations
    Research Center for Social Computing and Information Retrieval\\
    Harbin Institute of Technology, China
    \emails
    \{dzrwang, lxdou, car\}@ir.hit.edu.cn
}

\begin{document}
    \maketitle

    \begin{abstract}
        Table-and-text hybrid question answering (HybridQA) is a widely used and challenging NLP task commonly applied in the financial and scientific domain.
        The early research focuses on migrating other QA task methods to HybridQA, while with further research, more and more HybridQA-specific methods have been present.
        With the rapid development of HybridQA, the systematic survey is still under-explored to summarize the main techniques and advance further research.
        So we present this work to summarize the current HybridQA benchmarks and methods, then analyze the challenges and future directions of this task. 
        The contributions of this paper can be summarized in three folds: 
        (1) \tif{first survey}, to our best knowledge, including benchmarks, methods, and challenges for HybridQA;
        (2) \tif{systematic investigation} with the reasonable comparison of the existing systems to articulate their advantages and shortcomings; 
        (3) \tif{detailed analysis} of challenges in four important dimensions to shed light on future directions.
    \end{abstract}
    
    \section{Introduction}
        \label{Introduction}

The question-answering task with just text or tables to generate answers has been systematically studied, which we call the classic QA task.
These two sorts of evidence each have their advantages:
textual evidence is prevalent in daily communication, while tabular evidence is a well-organized display of numerical information. 
However, using the heterogeneous data that combines these two types of evidence is increasingly prevalent in real applications, particularly in fields demanding numerical reasoning, like the financial and scientific domains.
This technique is known as Table-and-Text Hybrid Question Answering (HybridQA).
To fill the gap of HybridQA research, an increasing number of benchmarks \cite{chen-etal-2020-hybridqa,zhu-etal-2021-tat} and associated methods have been present in recent years, drawing more and more attention to this task.
Considering that HybridQA is still under-researched and lacks a comprehensive survey, we present this paper to summarize the current development and help researchers get into this topic.

The HybridQA task requires the system to generate the answer to the question based on heterogeneous knowledge, including tables and text. 
Compared with the classic QA task, the HybridQA task requires the system to model these two types of evidence, which makes it harder to obtain the correct answers.

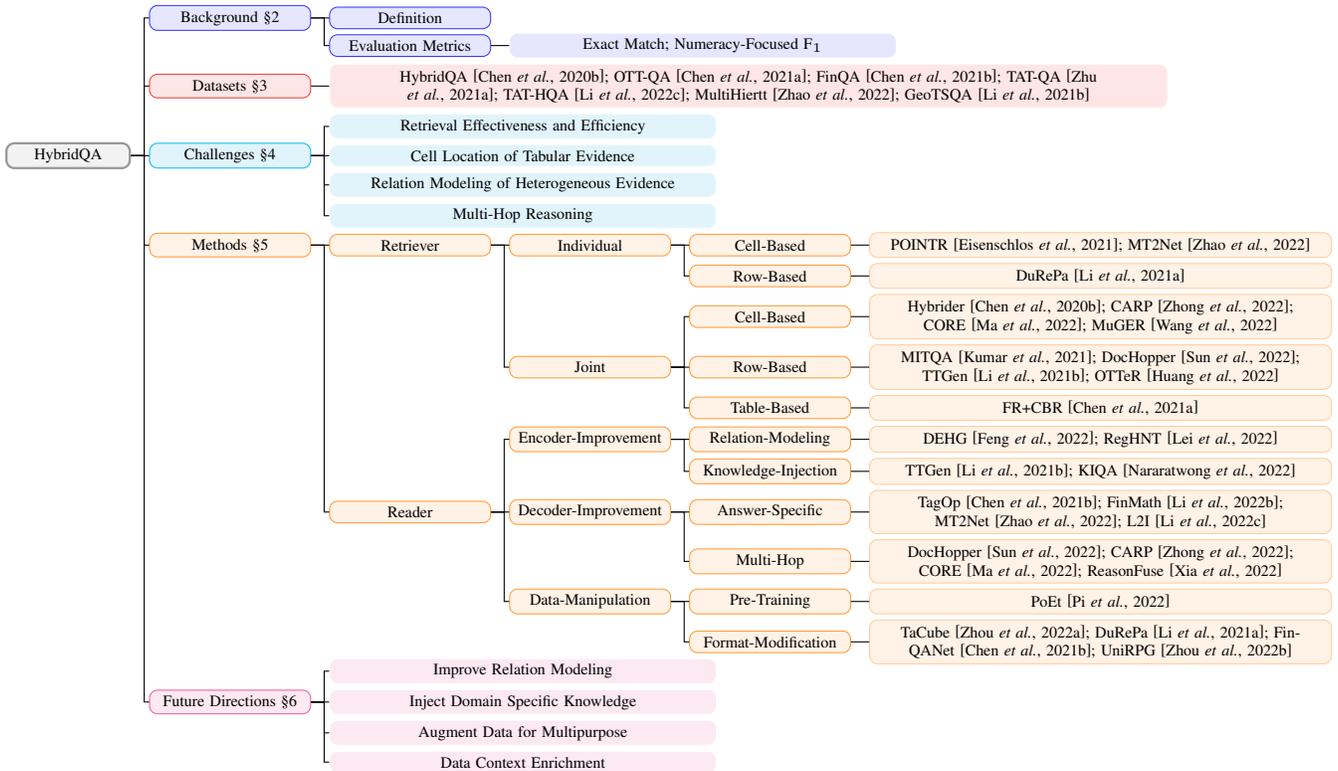
\begin{figure*}
    \tiny
    \begin{forest}
        for tree={
            forked edges,
            grow'=0,
            draw,
            rounded corners,
            s sep=2pt,
            calign=child edge, 
            calign child=(n_children()+1)/2
        },
        [HybridQA, parent
            [Background \S\ref{sec:background}, for tree={background}
                [Definition, background]
                [Evaluation Metrics, background
                    [Exact Match; Numeracy-Focused F$_1$, background_work]
                ]
            ]
            [Datasets \S\ref{sec:benchmarks}, for tree={datasets}
                [HybridQA \cite{chen-etal-2020-hybridqa}; OTT-QA \cite{chen2021open}; FinQA \cite{chen-etal-2021-finqa}; TAT-QA \cite{zhu-etal-2021-tat}; TAT-HQA \cite{li-etal-2022-learning}; MultiHiertt \cite{zhao-etal-2022-multihiertt}; GeoTSQA \cite{Li-teal-2021-tsqa}, datasets_work]
            ]
            [Challenges \S\ref{sec:challenges}, for tree={challenges}
                [Retrieval Effectiveness and Efficiency, challenges_work]
                [Cell Location of Tabular Evidence, challenges_work]
                [Relation Modeling of Heterogeneous Evidence, challenges_work]
                [Multi-Hop Reasoning, challenges_work]
            ]
            [Methods \S\ref{sec:methods}, for tree={methods}
                [Retriever, methods
                    [Individual, methods
                        [Cell-Based, methods
                            [POINTR \cite{eisenschlos-etal-2021-mate}; MT2Net \cite{zhao-etal-2022-multihiertt}, methods_work]
                        ]
                        [Row-Based, methods
                            [DuRePa \cite{li-etal-2021-dual}, methods_work]
                        ]
                    ]
                    [Joint, methods
                        [Cell-Based, methods
                            [Hybrider \cite{chen-etal-2020-hybridqa}; CARP \cite{zhong-etal-2022-reasoning}; CORE \cite{ma2022open}; MuGER \cite{wang2022muger}, methods_work]
                        ]
                        [Row-Based, methods
                            [MITQA \cite{kumar2021multi}; DocHopper \cite{sun2022iterative}; TTGen \cite{Li-teal-2021-tsqa}; OTTeR \cite{huang2022mixed}, methods_work]                            
                        ]
                        [Table-Based, methods
                            [FR+CBR \cite{chen2021open}, methods_work]
                        ]
                    ]
                ]
                [Reader, methods
                    [Encoder-Improvement, methods
                        [Relation-Modeling, methods
                            [DEHG \cite{feng-etal-2022-multi}; RegHNT \cite{lei-etal-2022-answering}, methods_work]
                        ]
                        [Knowledge-Injection, methods
                            [TTGen \cite{Li-teal-2021-tsqa}; KIQA \cite{nararatwong-etal-2022-kiqa}, methods_work]
                        ]
                    ]
                    [Decoder-Improvement, methods
                        [Answer-Specific, methods
                            [TagOp \cite{chen-etal-2021-finqa}; FinMath \cite{li-etal-2022-finmath}; MT2Net \cite{zhao-etal-2022-multihiertt}; L2I \cite{li-etal-2022-learning}, methods_work]
                        ]
                        [Multi-Hop, methods
                            [DocHopper \cite{sun2022iterative}; CARP \cite{zhong-etal-2022-reasoning}; CORE \cite{ma2022open}; ReasonFuse \cite{xia2022reasonfuse}, methods_work]
                        ]
                    ]
                    [Data-Manipulation, methods
                        [Pre-Training, methods
                            [PoEt \cite{pi2022reasoning}, methods_work]
                        ]
                        [Format-Modification, methods
                            [TaCube \cite{zhou2022tacube}; DuRePa \cite{li-etal-2021-dual}; FinQANet \cite{chen-etal-2021-finqa}; UniRPG \cite{zhou2022unirpg}, methods_work]
                        ]
                    ]
                ]
            ]
            [Future Directions \S\ref{sec:future}, future
                [Improve Relation Modeling, future_work]
                [Inject Domain Specific Knowledge, future_work]
                [Augment Data for Multipurpose, future_work]
                [Data Context Enrichment, future_work]
            ]
        ]
    \end{forest}
    \caption{Summary of HybridQA task.}
    \label{fig:summary}
    \vspace{-2.5em}
\end{figure*}

To advance these emerging and important research topics, several high-quality HybridQA benchmarks have been proposed. 
For example, HybridQA \cite{chen-etal-2020-hybridqa} mainly focuses on evidence extraction (i.e., finding the grounded truth from hundreds of evidence candidates).
In contrast, TAT-QA \cite{zhu-etal-2021-tat} concentrates on numerical reasoning (e.g., aggregation and sorting) in the hybrid context.

According to these benchmarks, we systematically summarize the challenges of HybridQA to deepen our understanding of the task, thereby inspiring more research ideas. 
Although different HybridQA benchmarks have significantly different settings, the core challenges of all benchmarks are the same, which make up the main challenges of the HybridQA task.
Concretely, we summarize four main challenges: retrieval effectiveness and efficiency, cell location of tabular evidence, relation modeling of heterogeneous evidence, and multi-hop reasoning.

To handle these challenges, current research introduces several effective methods.
Like classic QA systems with homogeneous evidence, the HybridQA system can also be divided into retriever and reader modules \cite{zhu2021retrieving}.
Most retrievers employ the pre-train language model (PLM) as the encoder, while the difference is the table retrieval granularity and whether to link heterogeneous evidence.
About the reader, some approaches use machine reading comprehension (MRC) \cite{chen-etal-2020-hybridqa,kumar2021multi}, while other systems are designed for HybridQA-specific challenges \cite{li-etal-2022-finmath,lei-etal-2022-answering}, which can be divided into encoder, decoder, and data manipulation.

Although the current methods have achieved remarkable improvement, they cannot completely solve all HybridQA challenges.
To shed light on HybridQA, we discuss several promising research directions in the future, which include relation-modeling improvement, specific-knowledge injection, data augmentation, and context enrichment. 
We hope these directions can make significant progress on the HybridQA task because they have been proven effective on many other tasks \cite{nakano2021webgpt,li-etal-2022-DASurvey}.

To summarize, our contributions include
\begin{itemize}[noitemsep, nolistsep, leftmargin=*]
    \item \textit{First Survey}:
        To the best of our knowledge, this is the first systematic survey about the HybridQA task with five parts of summarization (Figure~\ref{fig:summary}).
    \item \textit{Systematic Investigation}:
        We propose a reasonable comparison of the existing systems to articulate their advantages and shortcomings.
    \item \textit{Detailed Analysis}:
        We summarize the main challenges of HybridQA to deepen our understanding of this new task. Then we propose four promising future directions.
\end{itemize}

    \begin{table*}[ht]
        \centering
        \tiny
        \begin{tabular}{cc|cccccc|ccc}
            \toprule 
                \textbf{Context} & \textbf{Name} & \textbf{NR} & \textbf{Hypothesis} & \textbf{Hierarchical} & \textbf{Open-QA} & \textbf{Multi-Turn} & \textbf{Multi-Modal} & \textbf{Domain} & \textbf{\#Example} & \textbf{Answer Type} \\
            \midrule 
                \multirow{7}{*}{Table-and-Text}
                 & HybridQA \cite{chen-etal-2020-hybridqa}  & & & & & & &  Wikipedia & 69,611 & Spans \\
                 & OTT-QA \cite{chen2021open}  & & & & \Checkmark & & &  Wikipedia & 45,481 & Spans \\
                 & GeoTSQA \cite{Li-teal-2021-tsqa}  & \Checkmark & & & & & &  Geography & 1,012 & Choice \\
                 & FinQA \cite{chen-etal-2021-finqa}  & \Checkmark & & & & & &  Finance & 8,283 & DSL \\
                 & TAT-QA \cite{zhu-etal-2021-tat}  & \Checkmark & & few & & & &  Finance & 16,552 & Multi-Type \\
                 & TAT-HQA \cite{li-etal-2022-learning}  & \Checkmark & \Checkmark & few & & & &  Finance & 8,283 & Multi-Type \\
                 & MultiHiertt \cite{zhao-etal-2022-multihiertt}  & \Checkmark & & \Checkmark & & & &  Finance & 10,440 & Multi-Type \\
            \midrule
                \multirow{7}{*}{MISC}
                 & HybriDialogue \cite{nakamura-etal-2022-hybridialogue}  & & & & & \Checkmark & &  Wikipedia & 21,070 & Spans \\
                 & ConvFinQA \cite{chen-etal-2022-ConvFinQA}  & \Checkmark & & & & \Checkmark & &  Finance & 14,115 & DSL \\
                 & PACIFIC \cite{Deng2022PACIFICTP}  & \Checkmark & & & & \Checkmark & &  Finance & 19,008 & Multi-Type \\
                 & TAT-DQA \cite{zhu-etal-2022-towards}  & \Checkmark & & few & & & \Checkmark &  Finance & 16,558 & Multi-Type \\
                 & MMQA \cite{talmor2021multimodalqa}  & & & & & & \Checkmark &  Wikipedia & 29,918 & Spans \\
            \bottomrule
        \end{tabular}%
        \caption{
            \textbf{NR:} Whether the questions require numerical reasoning.
            \textbf{Hypothesis:} Whether the questions include the hypothesis.
            \textbf{Hierarchical:} Whether the tables are the hierarchical structure.
            \textbf{Open-QA:} Whether the benchmark is the open-QA benchmark.
            \textbf{Domain:} The domain of background information in the benchmark.
            \textbf{\#Example:} Number of examples.
            \textbf{Answer Type:} Covered answer types.
        }
        \label{tab:benchmark}
        \vspace{-1em}
    \end{table*}
    
    \section{Background}
        \label{sec:background}
        In this section, we will propose the task definition and introduce two widely used evaluation metrics.

\subsection{Task Definition}
    \begin{figure}[t]
    	\centering
    	\includegraphics[width=\linewidth]{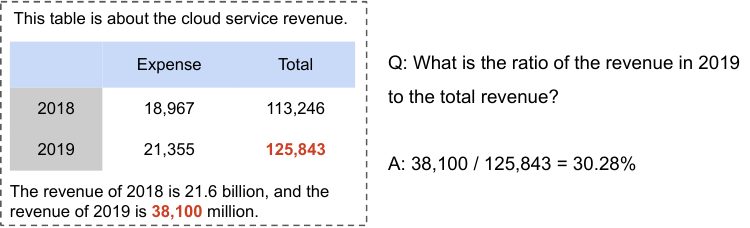}
    	\caption{
    	An example of HybridQA. 
            The question and answer (with an arithmetic formula) are shown on the right.
            The required evidence for answering this question is \textcolor{myRed}{\tf{bold}} on the left.
    	}
    	\label{fig:example}
        \vspace{-0.5em}
    \end{figure}

    The HybridQA model first takes a question as the input, retrieves the relevant tables and passages as the knowledge evidence, and generates a free-formed answer as the task output with reasoning over this retrieved evidence. 
    The type of answer includes 
    \ti{(1)} the single/multi-span, from either the table cells or the passages; 
    \ti{(2)} the calculation result, for arithmetic question; 
    \ti{(3)} the choice result, following the same set of the multi-choice QA task. 
    
    Each HybridQA data example $(Q, A, P, T, \hat{P}, \hat{T})$ consists of the question $Q$, the answer $A$, the passages and tables $(P, T)$, the corresponding textual and tabular evidences $(\hat{P}, \hat{T})$. 
    In some benchmarks, the answer $A$ is also affiliated with the arithmetic formula.

    In the case of Figure~\ref{fig:example}, given the question \tif{`What is the ratio of the revenue in 2019 to the total revenue?'}, the model first retrieves the relevant evidence, then generate formula \tif{`38,100 / 125,843'} by reasoning over the evidence, and finally calculates the answer \tif{`30.28\%'} based on the formula.

\subsection{Evaluation Metrics}
    The most common HybridQA evaluation metrics include
    \begin{itemize}[noitemsep, nolistsep, leftmargin=*]
        \item \textbf{Exact Match}  measures the percentage of predictions that match the ground truth answers. 
        Usually, two arithmetic answers are considered equal if their four decimal places are equal, following the rule of rounding function.
        \item \textbf{Numeracy-Focused F$_1$ Score} \cite{zhu-etal-2021-tat} measures the average token-level overlap between the predictions and the ground truth answers, which can reduce the false negative labeling.
        When an answer has multiple spans, the numeracy-focused F$_1$ performs a one-to-one alignment greedily based on the bag-of-word overlap on the set spans to ensure every current span can get the highest F$_1$ value, then compute micro-average F$_1$ over each span.
    \end{itemize}

    \section{Benchmarks}
        \label{sec:benchmarks}
        The HybridQA systems are driven by high-quality and large-quantity datasets. 
In this section, we introduce the widely-used HybridQA benchmarks. which are summarized in Table~\ref{tab:benchmark}.

\paragraph{HybridQA:}
    HybridQA \cite{chen-etal-2020-hybridqa} is the first HybridQA benchmark, which is also the largest cross-domain benchmark to date.
    Each question and answer is relayed on a single table and multiple texts.
    Each text usually is a description of information of a table cell, for example, a hyperlink page of the cell, which is crawled from Wikipedia. 
    For each case, the benchmark offers the golden text and table rows.
    All answers to questions are the spans in evidence, which called span-based answers, and need one or more hops between heterogeneous data.

\paragraph{OTT-QA:}
    To lower the difficulties of answering, HybridQA \cite{chen-etal-2020-hybridqa} annotates the related evidence to each example and the links of text and tables, which widens the gap with real-world applications.
    To be more relevant to the practical applications, OTT-QA \cite{chen2021open} blends textual and tabular evidence of each example into one single corpus that contains more than five million items and removes the relation information between them, which is called the open-QA benchmark. 
    So the most challenging part of this benchmark is to retrieve evidence of questions from millions of heterogeneous data, like open domain question answering. 
    The questions and evidence of OTT-QA are all built based on the HybridQA.
    Also, all its answers are the spans in the evidence.

\paragraph{FinQA:}
    Some HybridQA answers generation require numeric reasoning compatibility, while the benchmarks with only span-based questions cannot fulfill this requirement.
    FinQA \cite{chen-etal-2021-finqa} is a finance HybridQA benchmark containing the questions of many standard financial analysis calculations. 
    FinQA annotates the arithmetic answer in a domain-specific language (DSL), which consists of mathematical and table operations, to reduce the difficulty of formula generation and make it more interpretable.

\paragraph{TAT-QA:}
    Although FinQA has presented well-annotated numerical reasoning questions, it ignores the questions with span-based answers.
    Similar to the classic QA benchmark DROP \cite{dua-etal-2019-drop}, TAT-QA \cite{zhu-etal-2021-tat} is a collection of financial HybridQA samples that includes questions with both span-based and arithmetic answers.  
    Additionally, unlike the benchmarks mentioned above, each TAT-QA question is typically related to only five texts, which lowers the difficulty of retrieval. 
    Just like FinQA, TAT-QA also provides the formulations of arithmetic questions.

\paragraph{TAT-HQA:}
    In real applications, there exist many problems requiring hypothesis, for example, ``\textit{what is the balance of 2019 if the growth is the same as 2018?}".
    TAT-HQA \cite{li-etal-2022-learning} is a variant of TAT-QA \cite{chen-etal-2020-hybridqa} to simulate the hypothetical scenario questions by introducing assumptions based on the evidence. 
    The benchmark also annotates the hypothesis portion of each question in order to lessen the difficulty of model learning.

\paragraph{MultiHiertt:}
    Hierarchical tables, which contain multi-level headers, are common in the real world but are hard to be expressed and be understood by models because of the complex table structure. 
    However, almost all tables of the previous benchmarks are flattened structures without multi-level headers.
    To overcome this challenge, MultiHiertt \cite{zhao-etal-2022-multihiertt} collects and annotates many hierarchical tables compared with questions.

\paragraph{GeoTSQA:}
    GeoTSQA \cite{Li-teal-2021-tsqa} is the first scenario-based question-answering benchmark with hybrid evidence, which requires retrieving and integrating knowledge from multiple sources and applying general knowledge to a specific case described by the scenario.
    This benchmark is constructed on the multiple-choice questions in the geography domain from Chinese high-school exams.
    Besides tables and text, each question is also provided with four options, from which model should select one as the answer.

\paragraph{MISC:}
    To adapt the HybridQA task in more realistic applications compared with the benchmarks above, another group of benchmarks extends the vanilla HybridQA to more challenging scenarios,
    including the multi-turn \cite{nakamura-etal-2022-hybridialogue,chen-etal-2022-ConvFinQA,Deng2022PACIFICTP} and the multi-modal evidence \cite{zhu-etal-2022-towards,talmor2021multimodalqa}.
    However, since this survey mainly focuses on the HybridQA setting, we will not discuss these specific settings in the following. Please refer to the corresponding papers for more details as listed in~Table~\ref{tab:benchmark}.

    \begin{figure*}[t]
        \centering
        \includegraphics[width=\linewidth]{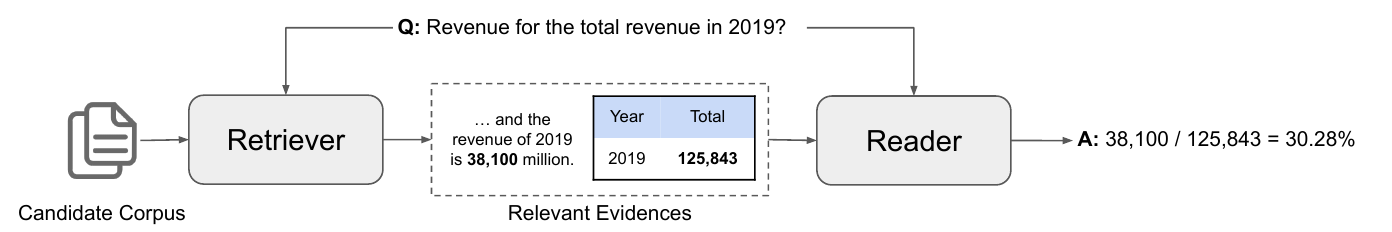}
        \caption{
            An illustration of the mainstream HybridQA system.
        }
        \label{fig:system}
        \vspace{-1em}
    \end{figure*}

    \section{Challenges}
        \label{sec:challenges}
        We can see from the datasets presented above that while having various settings, they all face similar difficulties, which are the main topics of the HybridQA approaches.
In the following, we identify four particular challenges of the HybridQA task and summarize them by illustrating \ti{(1) why is the challenge essential?} and \ti{(2) why is it so challenging?}

\paragraph{Trade-off between Retrieval Effectiveness and Efficiency.}
    The large-scale evidence could not be directly fed into the model due to limited input length. 
    Thus we should retrieve the most relevant evidence first.
    Effectiveness and efficiency are vital attributes in the retrieval problem \cite{karpukhin-etal-2020-dense}.
    Specifically, in the HybridQA problem, 
    (1) effectiveness (i.e., retrieval accuracy) would eventually determine the upper bound of QA accuracy and 
    (2) efficiency (i.e., retrieval latency) is essential for user experience in practical applications.
    However, these two attributes are difficult to optimize simultaneously. 
    It is because effective modeling usually needs deep semantic interaction and complex indexing, which would inevitably increase the cost of encoding.

\paragraph{Locating Tabular Evidence with Structure Modeling.}
    Tabular cell location requires the system to detect the cells related to the question.
    While it's tricky to understand the complex table structure.
    The models that could only capture the linear structure are difficulty encoding tables with complicated hierarchical structures.
    In addition, the question may not directly contain the related table header names, and the model needs to map the entities in the question to the correct cells based on semantics.

\paragraph{Relation Modeling of Heterogeneous Evidence.}
    The relation modeling identifies the relations of different evidence, which is the basis of many reasoning steps, such as multi-hop and numerical reasoning.
    The challenge is that the relations of the HybridQA task are more complex than the classic QA task, such as the positional relations in tables and the relations between tables and text.
    Thus the system should model the relations between heterogeneous evidence rightly and reduce the number of input relations that are not relevant to the question.
    In particular, two related pieces of evidence may only be semantically similar but not contain the same entity, which leads to the loss of effectiveness of traditional semantic matching methods.

\paragraph{Multi-Hop Reasoning.}
    Many real-world questions cannot be answered in one step but require multi-step reasoning on considerable evidence.
    The multi-hop reasoning is the method to solve this problem by obtaining the answer from different pieces of evidence based on the relations above.
    While the difficulty is that most multi-hop questions do not provide the solving process \cite{chen-etal-2021-finqa,zhu-etal-2021-tat}, which makes it hard to detect evidence reasoning order.
    Besides, the multi-hop reasoning question can be seen as the multi-turn questions mixed in one question, which requires the system to decompose it during encoding or to generate sub-questions during decoding.

    \section{Methods}
        \label{sec:methods}
        In this section, we introduce the recent progress of HybridQA in solving the challenges which are discussed in $\S$\ref{sec:challenges}. 
Following the definition of open-domain question answering \cite{zhu2021retrieving}, we divide the HybridQA system into two modules called Retriever ($\S$\ref{subsec:retriever}) and Reader ($\S$\ref{subsec:reader}).
The process of the HybridQA system can be summarized in Figure~\ref{fig:system}.
Notably, although some work \cite{kumar2021multi} build a unified framework to couple these two modules, to make it more clear, we still follow the mainstream work treating them as two separate modules.

\begin{table*}[t]
    \centering
    \small
    \begin{tabular}{cccc}
        \toprule 
         & \textbf{Cell-Based} & \textbf{Row-Based} & \textbf{Table-Based} \\
        \midrule 
        \multirow{2}{*}{\textbf{Individual}} & POINTR \cite{eisenschlos-etal-2021-mate} & \multirow{2}{*}{DuRePa \cite{li-etal-2021-dual}} &  \\
         & MT2Net \cite{zhao-etal-2022-multihiertt} &  &  \\
        \midrule 
        \multirow{4}{*}{\textbf{Joint}} & Hybrider \cite{chen-etal-2020-hybridqa} & MITQA \cite{kumar2021multi} & \multirow{4}{*}{FR+CBR \cite{chen2021open}} \\
         & CARP \cite{zhong-etal-2022-reasoning} & DocHopper \cite{sun2022iterative} &  \\
         & CORE \cite{ma2022open} & TTGen \cite{Li-teal-2021-tsqa} &  \\
         & MuGER \cite{wang2022muger} & OTTeR \cite{huang2022mixed} &  \\
        \bottomrule
    \end{tabular}
    \caption{
        Retrievers of HybridQA system.
    }
    \label{tab:retriever}
    \vspace{-1em}
\end{table*}

\subsection{Retriever}
    \label{subsec:retriever}
    The goal of the retriever is to find a set of evidence to induce the QA answer.
    Some HybridQA systems~\cite{eisenschlos-etal-2021-mate,chen-etal-2020-hybridqa,kumar2021multi,sun2022iterative} directly employ the two-stage retrieval methods from open-domain QA task (i.e., classic QA).
    They first adopt TF-IDF or BM25 \cite{robertson2009probabilistic} to filter out a fixed number of data~\ti{(coarse-grained ranking)}, then employ PLM to calculate the relevancy between the question and the filtered evidence~\cite{zhu2021retrieving}~\ti{(fine-grained ranking)}.
    The number of retrieval data in the first stage influences the trade-off between effectiveness and efficiency of retrieval \cite{chen2021open,zhong-etal-2022-reasoning,chen-etal-2020-hybridqa}.
    
    Besides these general retrievers that are directly inherited from the classic QA, many works advance the retriever to be HybridQA-specific for handling heterogeneous evidence.
    For instance, we could leverage 
    \ti{(1)} the relations between textual and tabular evidence and
    \ti{(2)} the tabular operation results just like SQL aggregate functions.
    In the following, we classify the current HybridQA retrievers from two aspects:
    \ti{(1)} the relations of the textual and tabular evidence, including Individually, Jointly and
    \ti{(2)} the table retrieval granularity, including Cell-Based, Row-Based, and Table-Based.
    The classification of retrievers about these two category methods is summarized in Table~\ref{tab:retriever}.
    Finally, we make a comparison of different types of retrievers.
    
    \subsubsection{5.1.1\ \ Individual}
        The individual retrievers consider textual and tabular evidence as two separate parts.
        This type of retriever can reduce the overhead of data processing and try to avoid cascading errors.
        \paragraph{Cell-Based:}
            Directly retrieving table cells is the simplest retrieval technique. 
            POINTR \cite{eisenschlos-etal-2021-mate} uses a creative table Transformer architecture to feed every table cell into the model. 
            However, this approach forces the model to figure out the relationships between various cells on its own, making learning more challenging. 
            To mitigate this problem, MT2Net \cite{zhao-etal-2022-multihiertt} models the cell relations by transferring each table into multiple sentences as the retrieval items, which describe the numerical relation of the table.
        \paragraph{Row-Based:}
            To maintain the table structure information (e.g., cells in the same row), utilizing rows as retrieval units is another method for maintaining cell relations. 
            For instance, DuRePa \cite{li-etal-2021-dual} linearizes and ranks every table row, where a specific token distinguishes cells of each row.

    \subsubsection{5.1.2\ \ Joint}
        The joint retrievers merge heterogeneous evidence as input.
        This kind of retriever takes into account the relations between the various kinds of evidence, and some of them even make these relations retrievable.
        \paragraph{Cell-Based:}
            Some benchmarks \cite{chen-etal-2020-hybridqa} have provided text and table relationships that the model can directly adapt. 
            Hybrider \cite{chen-etal-2020-hybridqa} connects cells with related text and question entities, and then every cell is scored and combined with that in the same row.
            Yet not all provided relations are useful, so MuGER \cite{wang2022muger} sees the linking also as the retrievable objects, just like cells or rows.
            To help models learn the process of reasoning, CARP \cite{zhong-etal-2022-reasoning}, and CORE \cite{ma2022open} search the relations between question entities, text entities, and table cells and try to select the golden linking path.
        \paragraph{Row-Based:}
            Just like DuRePa \cite{li-etal-2021-dual}, MITQA \cite{kumar2021multi} and DocHopper \cite{sun2022iterative} directly concatenate rows with related textual evidence as the retriever input.
            Taking into account that cell-based and row-based have different advantages, TTGen \cite{Li-teal-2021-tsqa} tries to mix two types of granularity to retrieve, which transfers table operations into sentences, such as extremum values and monotonic change, then ranks them with the question and related paragraphs.
            To recover the linking between tables and rows, OTTeR \cite{huang2022mixed} firstly links rows and text as blocks like OTT-QA baseline \cite{chen2021open}, then uses a modified dense retriever to obtain useful blocks.
        \paragraph{Table-Based:}
            Considering the prior knowledge that every text is only associated with one table, the Fusion Retriever and Cross-Block Reader system (FR+CBR), which is the baseline system of OTT-QA \cite{chen2021open}, presents the table-text group called block, which contains one table and several related texts. 
            All possible blocks will be flattened and retrieved to be the input of the following modules. 

    \subsubsection{5.1.3\ \ Comparison}
        \paragraph{Individual vs. Joint:}
            Because the relationships of different types of evidence can help the retriever comprehend heterogeneous knowledge, the joint retriever is adapted by the majority of current systems. 
            However, not all application scenarios and benchmarks provide the entities linked to the questions \cite{zhu-etal-2021-tat,chen-etal-2021-finqa}, and joint retrievers must establish the linking themselves.
            This could introduce incorrect information and result in a cascade of errors. 
            Individual retrievers work well on tiny-scale corpus benchmarks, which ask the reader to model relations instead of the retriever.
        \paragraph{Cell-Based vs. Row-Based vs. Table-Based:}
            The best is still debatable due to the benefits and drawbacks of various retrieval granularities. 
            Since HybridQA answers frequently incorporate information from single or multiple cells, cell-based retrievers can overlook a large amount of useless data. 
            However, they are unable to simulate the cross-row or cross-column information of the table structure. 
            The row-based retriever can better represent the row relation than the cell-based one, but it is still challenging to model the information about the column structure. 
            When there are too many words in a sequence, there will be a lot of useless information, and it will be difficult to add additional important information. 
            The only option to manage massive amounts of table data is to use a table-based retriever because it is difficult to divide each table into cells or rows.

\begin{table}[t]
    \centering
    \tablestyle{2.5pt}{1.0}
    \scalebox{0.7}{
        \begin{tabular}{lll}
            \toprule 
            \textbf{Reader} & \textbf{Method} & \textbf{System} \\
            \midrule 
            \multirow{4}{*}{Encoder-Improvement}
             & \multirow{2}{*}{Relation Modeling} & DEHG \cite{feng-etal-2022-multi} \\
             &   & RegHNT \cite{lei-etal-2022-answering} \\
            \cline{2-3} 
             & \multirow{2}{*}{Knowledge Injection} & TTGen \cite{Li-teal-2021-tsqa} \\
             &   & KIQA \cite{nararatwong-etal-2022-kiqa} \\
            \midrule 
            \multirow{7}{*}{Decoder-Improvement} 
             & \multirow{4}{*}{Multi-Tower Decoding} & TagOp \cite{zhu-etal-2021-tat} \\
             &   & FinMath \cite{li-etal-2022-finmath} \\
             &   & MT2Net \cite{zhao-etal-2022-multihiertt} \\
             &   & L2I \cite{li-etal-2022-learning} \\
            \cline{2-3} 
             & \multirow{5}{*}{Multi-Hop Reasoning} & DocHopper \cite{sun2022iterative} \\
             &   & CARP \cite{zhong-etal-2022-reasoning} \\
             &   & CORE \cite{ma2022open} \\
             &   & ReasonFuse \cite{xia2022reasonfuse} \\
             &   & ELASTIC \cite{zhang2022elastic} \\ 
            \midrule 
            \multirow{5}{*}{Data-Manipulation}
             & Pre-Training & PoEt \cite{pi2022reasoning} \\
            \cline{2-3} 
             & \multirow{4}{*}{Format Modification} & TaCube \cite{zhou2022tacube} \\
             &   & DuRePa \cite{li-etal-2021-dual} \\
             &   & FinQANet \cite{chen-etal-2021-finqa} \\
             &   & UniRPG \cite{zhou2022unirpg} \\
            \bottomrule
        \end{tabular}
    }
    \caption{
        Readers of HybridQA system.
    }
    \label{tab:reader}
    \vspace{-1em}
\end{table}

\subsection{Reader}
    \label{subsec:reader}

    Given the question and retrieved evidence, the role of the reader is to extract or generate the answer. 
    In the classic QA task, the systems use the MRC modules as their readers \cite{zhu2021retrieving}.
    Some HybridQA systems also employ the MRC modules as the readers for the span-based answers benchmarks \cite{chen-etal-2020-hybridqa,chen2021open,eisenschlos-etal-2021-mate,zhong-etal-2022-reasoning}.
    Many other systems have designed HybridQA-specific functions for the readers, such as the domain knowledge injection \cite{nararatwong-etal-2022-kiqa} and the relation modeling \cite{lei-etal-2022-answering}, which is the main topic of this section. 
    
\begin{figure}[t]
            \centering
            \includegraphics[width=0.7\linewidth]{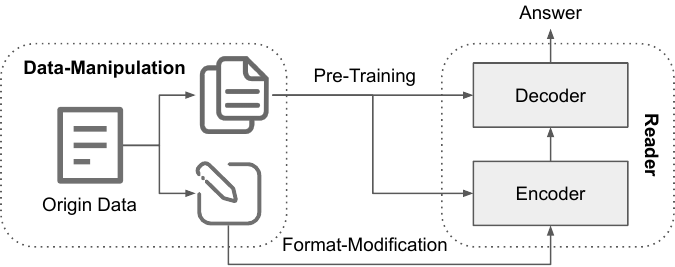}
            \caption{
                An illustration of the Data-Manipulation.
            }
            \label{fig:reader}
        \end{figure}

    In the following, we will introduce the current HybridQA system readers.
    We adopt the popular encoder-decoder framework to describe the current reader structure, where a series of works focused on improving the performance of the two modules.
    Besides, another series of work concentrates on the data instead of the model, which we called Data-Manipulation and summarized in Figure~\ref{fig:reader}.
    In summary, we categorize the improvement methods of the HybridQA reader into Encoder-Improvement, Decoder-Improvement, and Data-Manipulation.

    \subsubsection{5.2.1\ \ Encoder-Improvement}
        The encoder is the component that transforms the input into a format the model can understand. 
        Relation modeling and knowledge injection are just two of the techniques used in this section to increase the efficiency of encoders. 
        Knowledge injection focuses on how to teach the model with new information, while relation modeling focuses on how to construct the relationship of the evidence.
        \paragraph{Relation Modeling:}
            In contrast to traditional QA tasks, HybridQA data encoding requires the capacity to comprehend relationships.
            Thus DEHG \cite{feng-etal-2022-multi} introduces the relation graph to the HybridQA task, which first builds the graph of the evidence and the question according to the established rules, then linearizes the relation graph as an input. 
            However, DEHG is required to recover the graph by the linearized tuples on its own, which increases the difficulty of relation understanding.
            To better use graphs, RegHNT \cite{lei-etal-2022-answering} directly builds the graph network with model structure and adapts the relation-aware attention mechanism to encode different relation types.
        \paragraph{Knowledge Injection:}
            A specific domain leads to specific knowledge that may not be mentioned in training data.
            TTGen \cite{Li-teal-2021-tsqa} uses K-BERT \cite{liu-etal-2020-KBERT} to obtain the representation for each input token fused with the domain-specific knowledge, which is injected by the pre-training.
            Because the knowledge of Wikipedia is organized with entities (Wikipedia pages), KIQA \cite{nararatwong-etal-2022-kiqa} adapts more fine-grained knowledge injection, which firstly extracts entities from the question and evidence, then injects the knowledge about these entities into the model with the usage of LUKE \cite{yamada-etal-2020-luke}.

    \subsubsection{5.2.2\ \ Decoder-Improvement:}
        The decoder is the module to generate various types of answers to the questions based on the representation by the encoder.
        The methods of this part aim to improve the performance of the decoder, which includes multi-tower decoding and multi-hop reasoning.
        The multi-tower decoding focuses on designing different decoder structures for different answer types, while the latter improves the performance of multi-hop reasoning.
        \paragraph{Multi-Tower Decoding:}
            One of the challenges of HybridQA data decoding is generating widely different types of answers, such as the span-based and the arithmetic answers.
            TagOp \cite{zhu-etal-2021-tat} adopts a linear layer to decide the answer type of every example and determine which input tokens will be used, then generate the answer with the decoder corresponding to its type.
            Considering that the formula of arithmetic type answers can be represented in the tree format, FinMath \cite{li-etal-2022-finmath} uses Seq2Tree \cite{xie-etal-2019-goal} as the decoder to transfer a formula into a calculation tree, which is widely used in math word problem tasks.
            To address error cascades from the wrong type prediction, MT2Net \cite{zhao-etal-2022-multihiertt} predicts the probability of each type and span being the answer and selects the result with the highest joint probability as the final answer.
        \paragraph{Multi-Hop Reasoning:}
            Multi-hop is another decoder challenge, which requires the decoder to capture the chain of reasoning. 
            CARP \cite{pi2022reasoning}, and CORE \cite{ma2022open} build relation linking during retrieval, and the decoder needs to detect the linking path from question to answer spans. 
            The difference is that the latter uses a better method of generating and filtering relations.
            To simulate the process of humans solving multi-hop questions, DocHopper \cite{sun2022iterative} employs an iterative decoder that generates the question after one hop until obtaining the result.
            With the same motivation, ReansonFuse \cite{xia2022reasonfuse} adapts multiple LSTM decoding processes to ensure that the decoder holds encoded information and can use previous decoded results.
            Different from the above methods handle multiple types of multi-hop reasoning, ELASTIC \cite{zhang2022elastic} mainly focuses on arithmetic questions, which generate the operators and operands step by step.

    \subsubsection{5.2.3\ \ Data-Manipulation}
        
        Data manipulation, including pre-training and data format modification, is a series of simple but effective methods to enhance the table understanding of the model by adjusting the format and scale of the data without modifying the model structures.
        \paragraph{Pre-Training:}
            PoEt \cite{pi2022reasoning} adopts SQL execution task as a proxy of numerical reasoning task, considering the \ti{code execution} and \ti{language reasoning} share a similar protocol.
            For example, the model can learn the numerical reasoning capability by executing arithmetic formulations.
            Especially, PoEt first fine-tunes the pre-trained language model (e.g., BART and T5) on the synthesized code-related data, then fine-tunes it on the original HybridQA dataset.
        \paragraph{Format Modification:}
            Considering that the ability of PLM to understand data in various formats is different, modifying the input and output formats of the reader can significantly improve the ability to understand tables and generate answers. 
            TaCube \cite{zhou2022tacube} pre-generates information about tables, like the extremum value of every column called the cube.
            To avoid the embarrassment of table processing, DuRePa \cite{li-etal-2021-dual} generates SQL as well as answers and decides which one to use.
            Considering the poor numerical computing power of PLM, FinQANet \cite{chen-etal-2021-finqa} generates domain-specific language instead of the arithmetic results provided by FinQA. 
            While UniRPG \cite{zhou2022unirpg} extends numerical reasoning programs from tables to text, which can also use text spans as computation values.

    \section{Future Directions}
        \label{sec:future}
        As introduced above, the existing systems have almost addressed the main challenges ($\S$\ref{sec:challenges}).
Advanced techniques are being researched that could improve the current system by making it more powerful and reliable.
In the following, we will discuss these future directions by demonstrating \ti{(1) why are these techniques promising?} and \ti{(2) what is the tricky part in implementation?}

\paragraph{Improve Relation Modeling:}
    Relation modeling is the basis of functions such as evidence retrieval and multi-hop question answering in HybridQA tasks, but the current system still has many deficiencies.
    Although current methods have noticed different relations based on table structure \cite{lei-etal-2022-answering}, numerical comparisons and table header inclusion relations are still ignored.
    Most current systems use token-based methods to model evidence relations. 
    However, the neural network methods can extract relations with higher accuracy \cite{kolitsas-etal-2018-end}.
    With the relation modeling, current systems fed them into the retrievers and readers as the input. 
    At the same time, the relations can also be used to spread information through multiple iterations for multi-hop questions \cite{chen-etal-2020-question}.

\paragraph{Inject Domain Knowledge for Professional Experts:}
    In the realistic scenario, the HybridQA system should handle many domain-specific questions requiring specific knowledge.
    Injecting knowledge effectively mitigates the lack of domain-specific knowledge \cite{yamada-etal-2020-luke} of the HybridQA task.
    However, domain knowledge requires a specific format explaining entities, like a knowledge base or Wikipedia page, which need much human involvement.
    A possible solution is to use a text-based web browsing environment \cite{nakano2021webgpt}, and another solution is to transfer a document to the knowledge with the QA system \cite{Chen2022QAIT}.
    The current methods first train a module with injection knowledge \cite{nararatwong-etal-2022-kiqa,Li-teal-2021-tsqa}, while this approach may ignore some information because the model does not know which knowledge will be used in the actual Q\&A.
    So to make better use of the domain knowledge, after receiving a question, the model retrieves the knowledge related to the question, like a paragraph in the corpus or a node in the knowledge base, and then concatenates it after the question as the model inputs. 

\paragraph{Data-Augmentation for Handling Data-Scarcity:}
    The current scales of HybridQA benchmarks are not large because of the labeling difficulty, which limits the performance of the current systems.
    Data augmentation is an effective strategy to improve the performance of the system by automatically expanding the data scale \cite{li-etal-2022-DASurvey}, while it has not been systematically applied in the HybridQA task.
    A simple but effective augmentation method is to populate the templates with entities from the evidence \cite{yoran-etal-2022-turning}.
    Another method is to generate the question based on the given answer and evidence with the seq-to-seq model \cite{chen2021open}.
    The augmentation data can be directly used as training data.
    Considering the noise in this data, a more convenient method is to design the pre-training tasks for specific compatibility of HybridQA systems \cite{zhong-etal-2022-reasoning}.

\paragraph{Enrich Context Modeling for Realistic Scenario:}
    Most current research concentrates on simple context, i.e., single-turn questions and plain \ti{(text, table)} evidence.
    While in a realistic scenario, we could consider more complex context settings, such as multi-turn questioning and visually-rich evidence.
    Concretely, for multi-turn setting \cite{nakamura-etal-2022-hybridialogue,chen-etal-2022-ConvFinQA,Deng2022PACIFICTP}, it's more natural and informative in expressing the complex intent with follow-up questions rather than one single-turn question.
    But it's challenging to address the ellipsis and resolution problem in a multi-turn setting.
    For visually-rich evidence setting~\cite{talmor2021multimodalqa,zhu-etal-2022-towards}, the visual document exhibits the positional relation of \ti{(text, table)}.
    Intuitively, leveraging this type of visual relation would definitely improve the context modeling ability.
    Thus, it is worth exploring these promising context modeling for building an effective HybridQA model in realistic applications.

    \section{Conclusion}
        \label{sec:conclusion}
        In this paper, we comprehensively summarize and analyze the existing HybridQA task.
        Firstly, we study twelve mainstream HybridQA benchmarks, including their experiment settings and  application scenarios.
        Considering that the solutions of these benchmarks have strong commonalities, we analyze their most common challenges, which are also the main challenges of the HybridQA task.
        To demonstrate the current progress in addressing these challenges, we outline the present methods for them, compared with their advantages and disadvantages.
        Although current approaches have improved systems in many aspects, improvements in some parts are still very insufficient.
        So at the end of this paper, we propose four important but under-explored directions of the HybridQA task to inspire a more robust and reliable HybridQA system.

    \newpage
    
    \bibliographystyle{named}
    \bibliography{ijcai23}
\end{document}